\documentclass[10pt,twocolumn,letterpaper]{article}
\usepackage{multirow}

\usepackage{iccv}              %

\newcommand{\red}[1]{{\color{red}#1}}

\definecolor{iccvblue}{rgb}{0.21,0.49,0.74}
\usepackage[pagebackref,breaklinks,colorlinks,allcolors=iccvblue]{hyperref}

\def\paperID{40} %
\def\confName{MMFM2025}
\def\confYear{2025}

\title{RePOPE: Impact of Annotation Errors on the POPE Benchmark}

\author{ 
  \begin{tabular}{c}
    Yannic Neuhaus \quad\quad Matthias Hein \\\\
    T\"ubingen AI Center – University of T\"ubingen\\
  \end{tabular}
}

\begin{document}
\maketitle

\begin{abstract}
    Since data annotation is costly, benchmark datasets often incorporate labels from established image datasets. In this work, we assess the impact of label errors in MSCOCO on the frequently used object hallucination benchmark POPE. We re-annotate the benchmark images and identify an imbalance in annotation errors across different subsets. Evaluating multiple models on the revised labels, which we denote as RePOPE, we observe notable shifts in model rankings, highlighting the impact of label quality. Code and data are available at \url{https://github.com/YanNeu/RePOPE}.
\end{abstract}

\vspace{-4mm}

\section{Introduction}
The POPE~\cite{li2023evaluatingobjecthallucinationlarge} dataset has become a standard benchmark for object hallucinations in vision large language models (VLMs) and is frequently used by the research community, e.g. as part of the OpenVLM Leaderboard~\cite{duan2024vlmevalkit}. The most commonly used version of the POPE benchmark relies on the MSCOCO~\cite{lin2015microsoftcococommonobjects} image dataset which provides exhaustive annotations for 80 different object classes. It is known that image datasets such as MSCOCO contain significant amounts of annotation errors~\cite{schubert2023identifyinglabelerrorsobject}. In this work, we identify these errors and examine how they influence the results of the POPE benchmark by evaluating a corrected version which we denote as RePOPE. Our contributions are:
\begin{itemize}
    \item We assess annotation quality for the MSCOCO images used in the POPE benchmark.
    \item We provide RePOPE, a corrected label set for the benchmark, and show that the errors significantly impact the results.
\end{itemize}

\section{POPE}

POPE \cite{li2023evaluatingobjecthallucinationlarge} evaluates object hallucinations as a binary classification task, prompting VLMs with the question ``Is there a \textlangle\textit{object}\textrangle~in the image?''. The most common variant of POPE is based on 500 randomly selected images from the validation set of MSCOCO~\cite{lin2015microsoftcococommonobjects} which contain at least 3 objects according to the annotations. For each image, 6 questions are constructed, three with ground truth ``Yes'' and three with ground truth ``No''. While the three ``Yes''-questions can be directly derived from the MSCOCO annotations, questions with answer ``No'' are built by sampling from the non-annotated objects for the corresponding image. Note that all 80 MSCOCO object classes were exhaustively annotated for all images, i.e. all objects that are not annotated  can be assumed to be not present in the image. There are three sampling strategies proposed in \cite{li2023evaluatingobjecthallucinationlarge}, resulting in 3 variants of the benchmark. Note that all three variants share the same images as well as the same set of questions with answer ``Yes'', and differ only in the set of questions with ``No''. The three strategies for sampling those objects are:

\begin{itemize}
    \item \textbf{Random Sampling}: 3 objects are randomly sampled from all objects that are not annotated for the image
    \item \textbf{Popular Sampling}: the 3 most frequent objects in the image dataset which are not annotated for this image
    \item \textbf{Adversarial Sampling}: the 3 objects co-occuring most frequently with the objects that are actually present in the image 
\end{itemize}

In total, POPE consists of 3 sets of image-question pairs, each containing 1500 pairs with answer ``Yes'' and 1500 pairs with answer ``No''.

\section{RePOPE}
The construction of POPE relies on the original MSCOCO annotations. We re-annotate all 500 images and assign the labels:

\begin{itemize}
    \item ``Yes'' if the object is visible in the image,
    \item ``No'' if the object is \textbf{not} visible in the image,
    \item ``Ambiguous'' for corner cases where it is not clear whether the object is present or not,
\end{itemize}
based on consensus decision of two human labelers.

\begin{figure*}[h]
    \centering

    \begin{tabular}{c}
        \includegraphics[width=\linewidth]{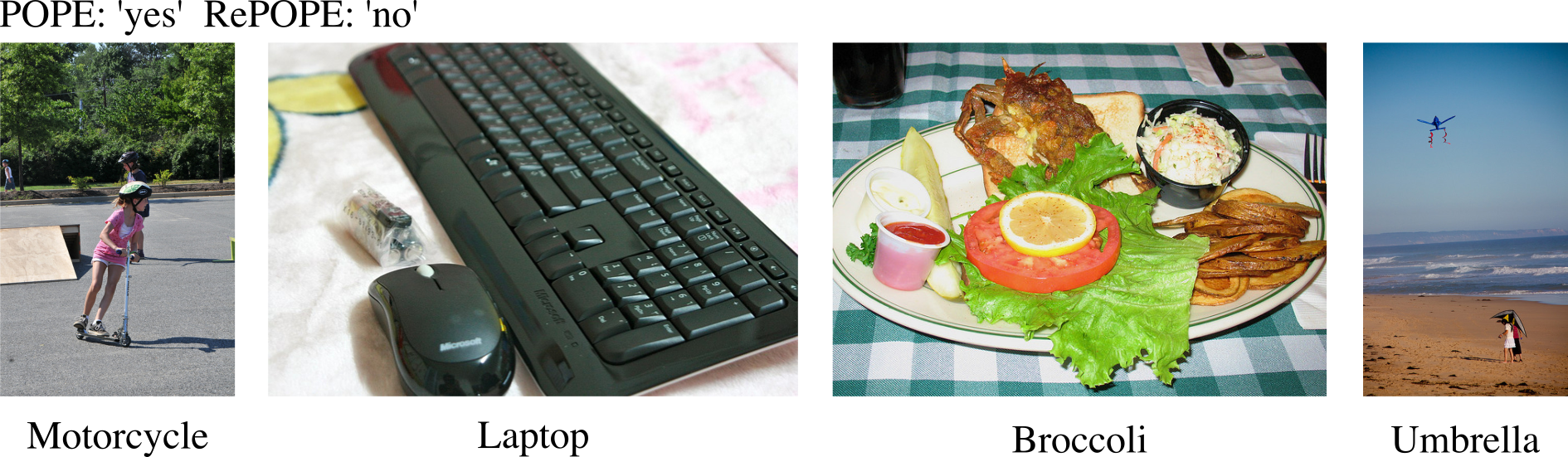}\\
        \midrule
        \includegraphics[width=\linewidth]{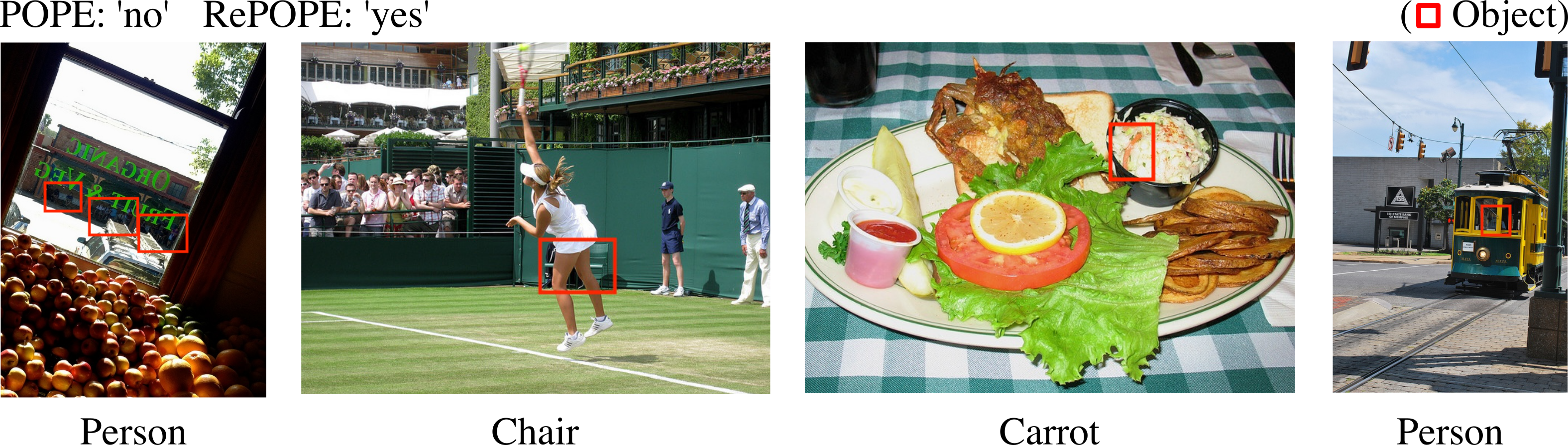}\\
        \midrule
        \includegraphics[width=\linewidth]{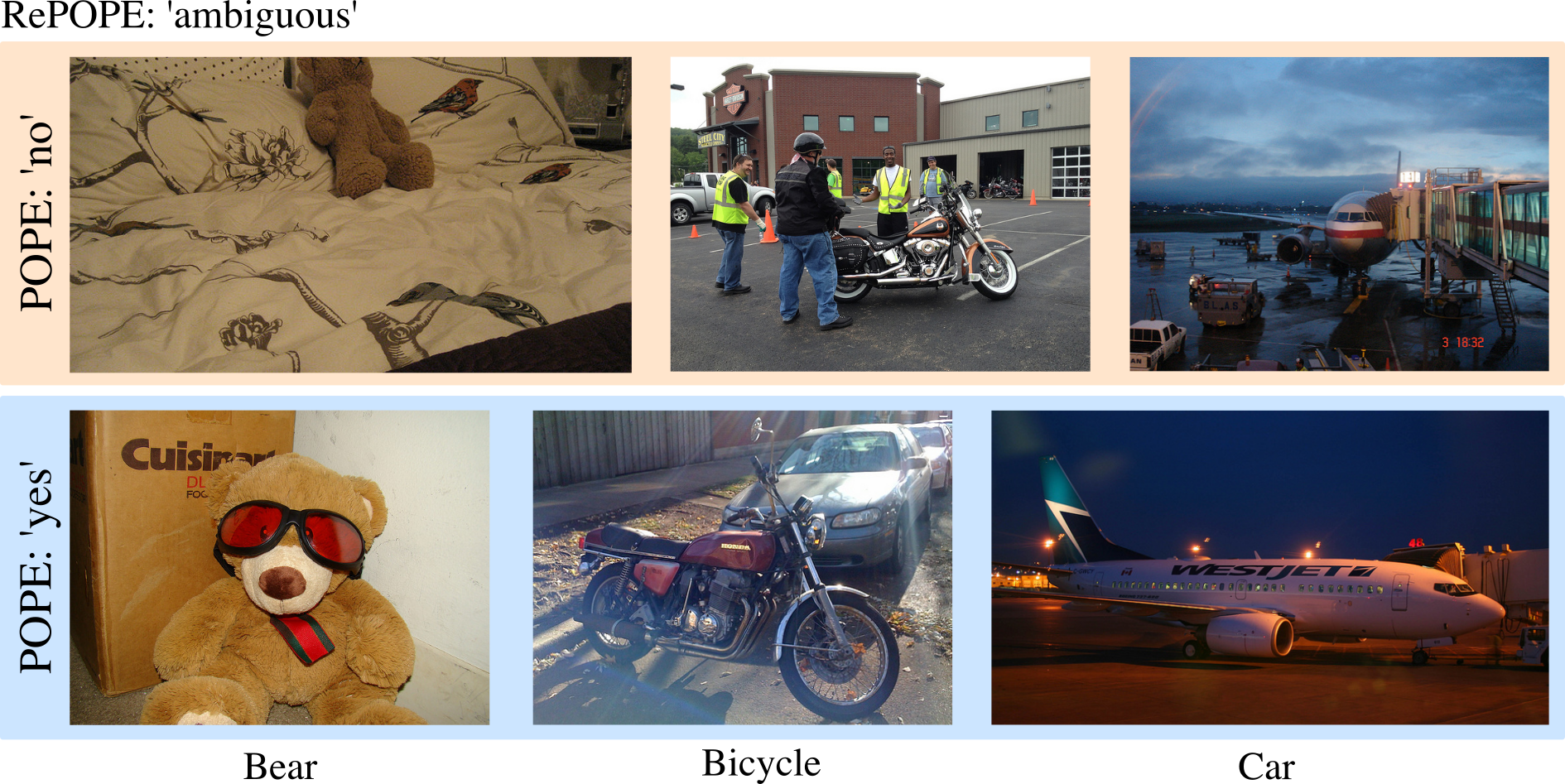}
    \end{tabular}

    \caption{\textbf{RePOPE annotation examples} The first row displays images that do not contain the object but are incorrectly labeled as “Yes” in POPE. The second row shows images where the object is present but mistakenly labeled as “No.” The object's presence is highlighted with a red box. The third row illustrates cases of inconsistent labeling in POPE, which we annotate as “ambiguous.” Examples include a “teddy bear” being categorized as a bear, a motorcycle being considered a motorized bicycle, and airport vehicles being classified as cars. Since these categorizations are subjective and MSCOCO labels are inconsistent, we exclude such cases from the benchmark.}
    \label{fig:label-examples}
\end{figure*}

\noindent Fig.~\ref{fig:label-examples} presents examples for our re-labeling in . For the positive set of POPE, i.e. the images with ground truth ``Yes'', most of the observed label errors are due to the presence of a visually similar or related object, e.g. a scooter mistaken for a ``motorcycle'', a mouse and keyboard labeled as ``laptop'', parsley on a plate as ``broccoli'', and a kite carried over the shoulder as an ``umbrella''. On the other hand, annotation errors on the negative set, i.e. images with ground truth ``No'', occur due to the very subtle presence of the object. The original annotators missed persons in the background or behind glass, the tennis player occludes the ``chairs'' in the background and the cole slaw contains only a small visible stripe of a carrot. For some objects, the COCO annotations are highly inconsistent likely due to differing definitions of those objects used by the original annotators. The classification of a ``teddy bear'' as a ``bear'', a motorcycle as a motorized ``bicycle'', or an airport vehicle as a ``car'' depends on specific definitions, leading to inconsistencies in POPE ground truth annotations. Therefore, we annotate the corresponding image-question pairs as ``ambiguous''.
Tab.~\ref{tab:relabeling_results} presents the results of our re-labeling. We observe a much higher error rate for the positive questions, i.e. the ones with answer ``Yes'', with $9.3\%$ annotation errors and $13.8\%$ ambiguous labels. In contrast, negative questions show a lower error rate, with $1.7\%$ labeling errors and $4.3\%$ ambiguous labels. The increasing error rate across the three subsets aligns with the expected occurrence frequency in the benchmark design where 'random' includes randomly selected objects, 'popular' consists of frequently occurring objects, and 'adversarial' features objects that frequently co-occur. 

\begin{table}[h]
    \centering
    \small
    \setlength{\tabcolsep}{3pt} %
    \begin{tabular}{l|ccc|ccc}
        \toprule
         & \multicolumn{3}{c|}{\textbf{POPE: Yes}} & \multicolumn{3}{c}{\textbf{POPE: No}} \\
        \cmidrule(lr){2-4} \cmidrule(lr){5-7}
        RePOPE Labels& Yes & \red{No} & Amb. & \red{Yes} & No & Amb. \\
        \midrule
        Random & \multirow{3}{*}{$76.9\%$} & \multirow{3}{*}{\red{$9.3\%$}} & \multirow{3}{*}{$13.8\%$} 
        & $\red{0.3\%}$ & $98.4\%$ & $1.3\%$ \\
        Popular &  & &  
        & $\red{2.6\%}$ & $93.0\%$ & $4.4\%$ \\
        Adversarial &  &  &   
        & $\red{2.2\%}$ & $90.5\%$ & $7.3\%$ \\
        \midrule
        Total& $76.9\%$ & $\red{9.3\%}$ & $13.8\%$ 
        & $\red{1.7\%}$ & $94.0\%$ & $4.3\%$\\
        \bottomrule
    \end{tabular}
    \caption{\textbf{Results of the re-annotation} The positive questions are identical for all three POPE variants. Among the questions with POPE answer ``Yes'', we find $9.3\%$ label errors and $13.8\%$ ambiguous cases where the correct label is not clear. For the questions with POPE answer ``No'', only $1.7\%$ of the questions have a wrong label and $4.3\%$ are ambiguous.}
    \label{tab:relabeling_results}
\end{table}

\section{Experiments}
For our corrected benchmark RePOPE, we correct all ground truth labels where the original annotations disagree with our re-labeling (yes/no or no/yes) and remove all image-question pairs that were annotated as ambiguous. We evaluate models on both label sets and compare the resulting metrics, either considering the values on the individual splits (random, popular, adversarial) or the mean over all three (mean).

\subsection{Models}
We evaluate a range of open-weight models on POPE and RePOPE covering different architectures and model sizes. Included are also some of the top models for POPE on OpenVLM Leaderboard~\cite{duan2024vlmevalkit}: InternVL2.5~\cite{chen2024expanding} (8B/26B/38B/78B and 8B-MPO/26B-MPO), LLaVA-NeXT~\cite{liu2024llavanext} (Vicuna\cite{peng2023vicuna}/Mistral\cite{jiang2023mistral7b}/Llama\cite{dubey2024llama3herdmodels}), LLaVA-OneVision~\cite{li2024llavaOneVision}, Ovis2~\cite{lu2024ovis} (1B/2B/4B/8B), PaliGemma-3B~\cite{beyer2024paligemma} and PaliGemma2~\cite{steiner2024paligemma} (3B/10B).

\subsection{Results}
\begin{figure*}
    \centering

    \begin{tabular}{c}
        True Positives (TP) and False Positives (FP)\\
        \includegraphics[width=\linewidth]{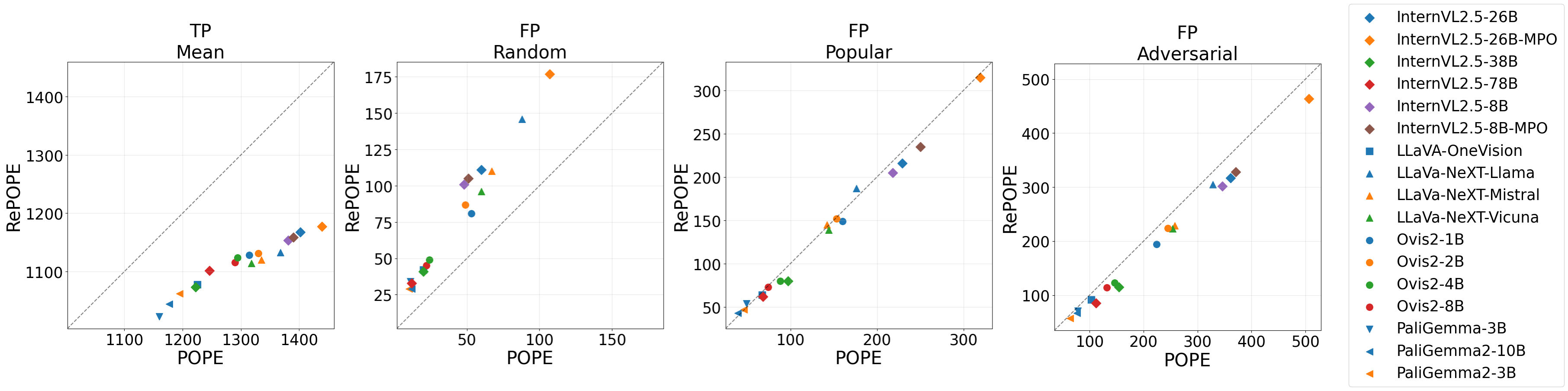} \\
        \midrule
        F1 Scores\\
        \includegraphics[width=\linewidth]{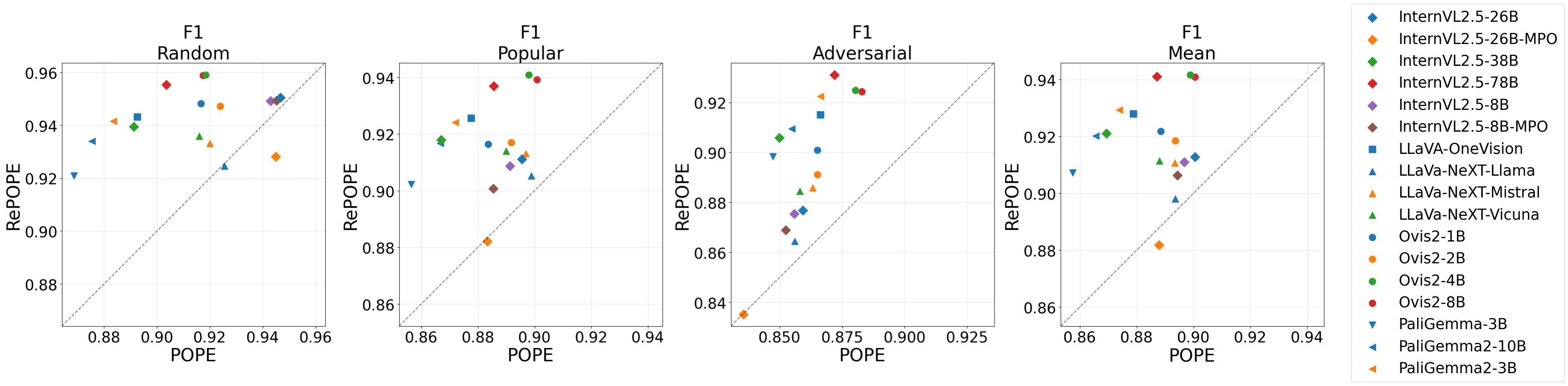}
    \end{tabular}
    \caption{\textbf{POPE vs RePOPE:} Due to the high error rate on the positive labels, the number of TP is significantly reduced across all models. Regarding FP, we observe different patterns across the three subsets: on the random subset, the number of false positives almost doubles for most of the models, results on popular are relatively stable, and on adversarial we observe a slight reduction of false positives. The ranking according to the F1 score is heavily impacted by the relabeling. The top models (Ovis2-4B/-8B) on the popular and adversarial split for POPE, also achieve the top ranks on random for RePOPE.}
    \label{fig:scatter}
\end{figure*}

In this section, we investigate how the biased distribution of label errors impacts the results on the POPE benchmark. In Fig.~\ref{fig:scatter}, we show how the number of true positives (TP) and false positives (FP) changes after the relabeling (as there is almost no variance across the positve image-question pairs of the three variants, we only report the mean over all three for TPs). While TP counts drop significantly, FP changes follow a more nuanced pattern. On the random subset, the number of FP almost doubles for most models, i.e. half of the objects that are falsely recognized by the models are covered by annotation errors on this POPE variant. This questions how reliable this kind of error can be measured on this split and suggests that the negative set is saturated on POPE random. For the adversarial variant, the number of FPs even decreases, most likely due to a higher prevalence of label errors on the negative set, i.e. by selecting images of frequently co-occurring objects it gets more likely that the object of interest is also in the image. Note that the rankings are relatively stable considering these counts. This is also true when considering precision and recall. In general, the models show an improved recall on RePOPE while their precision decreases but the rankings stay roughly similar for both metrics and reflect the ranking according to the yes ratio (see Fig.~\ref{fig:tpr-tnr-scatter}). Nevertheless, the relative shifts significantly impact the ranking according to F1 scores (POPE's main metric) as shown in Fig.~\ref{fig:scatter}. On the random subset, the top models of the RePOPE ranking, Ovis2-4B and -8B, are aligned with the top models for both POPE and RePOPE on the popular and adversarial subsets, indicating that the larger number of false positives in the subset enables a better measurement of hallucinations. Some models that achieve some of the best F1 scores in the POPE ranking, e.g. InternVL2.5-8B or -26B, drop to the bottom of the ranking after evaluating on the RePOPE labels. A similar pattern holds for the accuracy scores (as shown in Fig.~\ref{fig:acc-scatter}). However, as the corrected labels are not balanced anymore with respect to the amount of positive and negative samples, acccuracy values on RePOPE might be biased.
We provide full results tables for POPE (App.~\ref{app:full-pope}) and RePOPE (App.~\ref{app:full-repope}) in the appendix.

\section{Conclusion}
In this work, we explored the impact of annotation errors in the MSCOCO image dataset on the results of the POPE object hallucination benchmark. We observe a substantial larger amount of label errors on the positive set of POPE (answer ``Yes'') compared to the negative set (answer ``No'') which translates into a significant change in the F1 score rankings after re-labeling the images. This significant influence of the identified annotation errors on the benchmark results highlights the importance of data quality. To enable a more robust measurement of vulnerability to object hallucinations, we provide the corrected labels under the name RePOPE. Note that this re-labeling has only a limited effect on the saturation of the benchmark (many models acchieve true positive rates as well as true negative rates of more than $90\%$. To overcome this, other benchmarks need to be evaluated complimentary, e.g. DASH-B~\cite{augustin2025dashdetectionassessmentsystematic} which follows a similar design as POPE with a ``harder'' negative set.

{
    \small
    \bibliographystyle{ieeenat_fullname}
    \bibliography{main}

\begin{thebibliography}{14}
\providecommand{\natexlab}[1]{#1}
\providecommand{\url}[1]{\texttt{#1}}
\expandafter\ifx\csname urlstyle\endcsname\relax
  \providecommand{\doi}[1]{doi: #1}\else
  \providecommand{\doi}{doi: \begingroup \urlstyle{rm}\Url}\fi

\bibitem[et~al.(2025)]{augustin2025dashdetectionassessmentsystematic}
Augustin et al.
\newblock Dash: Detection and assessment of systematic hallucinations of vlms.
\newblock \emph{arXiv preprint arXiv:2503.23573}, 2025.

\bibitem[et~al.(2024{\natexlab{a}})]{beyer2024paligemma}
Beyer et al.
\newblock Pali{G}emma: A versatile {3B VLM} for transfer, 2024{\natexlab{a}}.

\bibitem[et~al.(2024{\natexlab{b}})]{chen2024expanding}
Chen et al.
\newblock Expanding performance boundaries of open-source multimodal models with model, data, and test-time scaling.
\newblock \emph{arXiv preprint arXiv:2412.05271}, 2024{\natexlab{b}}.

\bibitem[et~al.(2024{\natexlab{c}})]{duan2024vlmevalkit}
Duan et al.
\newblock Vlmevalkit: An open-source toolkit for evaluating large multi-modality models.
\newblock In \emph{Proceedings of the 32nd ACM International Conference on Multimedia}, pages 11198--11201, 2024{\natexlab{c}}.

\bibitem[et~al.(2024{\natexlab{d}})]{dubey2024llama3herdmodels}
Dubey et al.
\newblock The llama 3 herd of models, 2024{\natexlab{d}}.

\bibitem[et~al.(2023{\natexlab{a}})]{jiang2023mistral7b}
Jiang et al.
\newblock Mistral 7b, 2023{\natexlab{a}}.

\bibitem[et~al.(2023{\natexlab{b}})]{li2023evaluatingobjecthallucinationlarge}
Li et al.
\newblock Evaluating object hallucination in large vision-language models, 2023{\natexlab{b}}.

\bibitem[et~al.(2024{\natexlab{e}})]{li2024llavaOneVision}
Li et al.
\newblock Llava-onevision: Easy visual task transfer.
\newblock \emph{arXiv preprint arXiv:2408.03326}, 2024{\natexlab{e}}.

\bibitem[et~al.(2024{\natexlab{f}})]{liu2024llavanext}
Liu et al.
\newblock Llava-next: Improved reasoning, ocr, and world knowledge, 2024{\natexlab{f}}.

\bibitem[et~al.(2024{\natexlab{g}})]{lu2024ovis}
Lu et al.
\newblock Ovis: Structural embedding alignment for multimodal large language model.
\newblock \emph{arXiv preprint arXiv:2405.20797}, 2024{\natexlab{g}}.

\bibitem[et~al.(2023{\natexlab{c}})]{peng2023vicuna}
Peng et al.
\newblock Instruction tuning with gpt-4, 2023{\natexlab{c}}.

\bibitem[et~al.(2023{\natexlab{d}})]{schubert2023identifyinglabelerrorsobject}
Schubert et al.
\newblock Identifying label errors in object detection datasets by loss inspection.
\newblock \emph{arXiv preprint arXiv:2303.06999}, 2023{\natexlab{d}}.

\bibitem[et~al.(2024{\natexlab{h}})]{steiner2024paligemma}
Steiner et al.
\newblock Paligemma 2: A family of versatile vlms for transfer.
\newblock \emph{arXiv preprint arXiv:2412.03555}, 2024{\natexlab{h}}.

\bibitem[et~al.(2015)]{lin2015microsoftcococommonobjects}
Tsung-Yi et al.
\newblock Microsoft coco: Common objects in context, 2015.

\end{thebibliography}
}

\clearpage
\appendix

\section{Appendix: Additional Plots}
\label{app:add-plots}
\begin{figure*}[h]
    \centering
    \includegraphics[width=\linewidth]{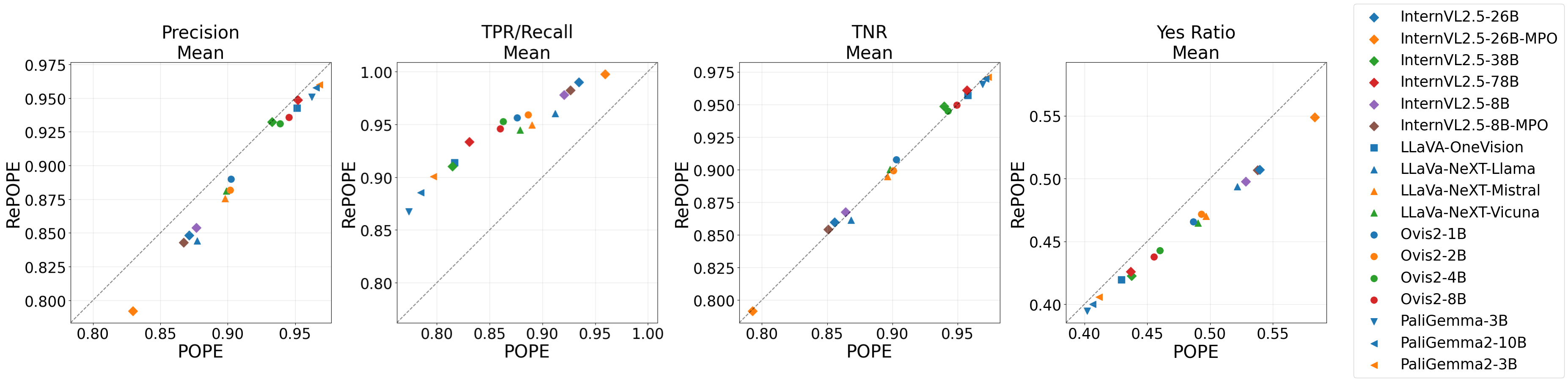}
    \caption{\textbf{POPE vs. RePOPE: Precision and Recall} The models' precision decreases on RePOPE while the true positive rate (TPR) improves. Effects on the true negative rate (TNR) are small but sufficient to change the ranking, and due to the larger amount of label errors in the positive questions, the models' yes rates decrease on the relabeled dataset.}
    \label{fig:tpr-tnr-scatter}
\end{figure*}

\begin{figure*}[h]
    \centering
    \includegraphics[width=\linewidth]{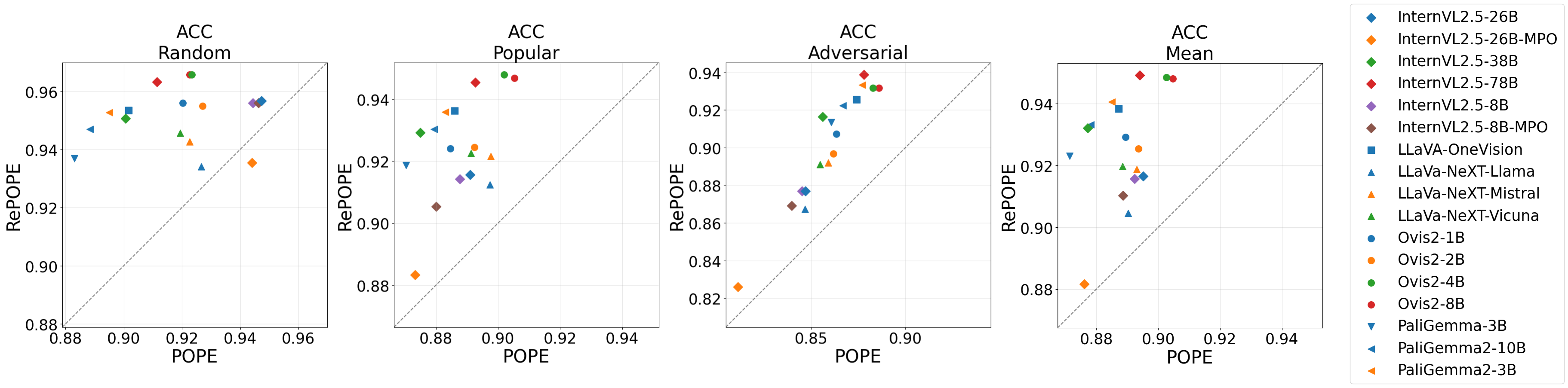}
    \caption{\textbf{POPE vs. RePOPE: Accuracy} We observe a similar pattern as for the F1 score. However, note that for RePOPE the number of positive and negative samples is not balanced anymore. Thus, accuracy needs to be interpreted with care.}
    \label{fig:acc-scatter}
\end{figure*}

We show additional scatter plots including precision, recall, true negative rates and yes ratios in Fig.~\ref{fig:tpr-tnr-scatter} as well as accuracy over all POPE subsets in Fig.~\ref{fig:acc-scatter}.

\section{Appendix: Full Results POPE}
\label{app:full-pope}

We show all results for all models on the random (Tab.~\ref{tab:POPE-random}), popular (Tab.~\ref{tab:POPE-popular}), and adversarial (Tab.~\ref{tab:POPE-adversarial}) split as well as the mean over all three subsets (Tab.~\ref{tab:POPE-Mean}).

\begin{table*}[]
    \centering
    \begin{tabular}{rl|c|c|c|c|c|c|c|c|c}
        \toprule
        Pos. & Model & F1 & TP & FP & TN & FN & Precision & Recall & ACC & Yes Ratio\\
        \midrule
        1 & InternVL2.5-26B & $90.1\%$ & 1402.0 & 216.7 & 1283.3 & 98.0 & $87.1\%$ & $93.5\%$ & $89.5\%$ & $54.0\%$ \\
        2 & Ovis2-8B & $90.0\%$ & 1290.0 & 76.0 & 1424.0 & 210.0 & $94.5\%$ & $86.0\%$ & $90.5\%$ & $45.5\%$ \\
        3 & Ovis2-4B & $89.9\%$ & 1294.0 & 86.0 & 1414.0 & 206.0 & $93.9\%$ & $86.3\%$ & $90.3\%$ & $46.0\%$ \\
        4 & InternVL2.5-8B & $89.7\%$ & 1381.0 & 204.0 & 1296.0 & 119.0 & $87.7\%$ & $92.1\%$ & $89.2\%$ & $52.8\%$ \\
        5 & InternVL2.5-8B-MPO & $89.4\%$ & 1390.0 & 224.0 & 1276.0 & 110.0 & $86.7\%$ & $92.7\%$ & $88.9\%$ & $53.8\%$ \\
        6 & Ovis2-2B & $89.4\%$ & 1330.0 & 149.0 & 1351.0 & 170.0 & $90.2\%$ & $88.7\%$ & $89.4\%$ & $49.3\%$ \\
        7 & LLaVa-NeXT-Llama & $89.3\%$ & 1368.0 & 197.3 & 1302.7 & 132.0 & $87.7\%$ & $91.2\%$ & $89.0\%$ & $52.2\%$ \\
        8 & LLaVa-NeXT-Mistral & $89.3\%$ & 1335.0 & 155.7 & 1344.3 & 165.0 & $89.8\%$ & $89.0\%$ & $89.3\%$ & $49.7\%$ \\
        9 & Ovis2-1B & $88.8\%$ & 1314.0 & 145.7 & 1354.3 & 186.0 & $90.2\%$ & $87.6\%$ & $88.9\%$ & $48.7\%$ \\
        10 & LLaVa-NeXT-Vicuna & $88.8\%$ & 1318.0 & 152.7 & 1347.0 & 182.0 & $89.9\%$ & $87.9\%$ & $88.8\%$ & $49.0\%$ \\
        11 & InternVL2.5-26B-MPO & $88.8\%$ & 1439.0 & 310.7 & 1189.3 & 61.0 & $83.0\%$ & $95.9\%$ & $87.6\%$ & $58.3\%$ \\
        12 & InternVL2.5-78B & $88.7\%$ & 1246.0 & 64.0 & 1436.0 & 254.0 & $95.2\%$ & $83.1\%$ & $89.4\%$ & $43.7\%$ \\
        13 & LLaVA-OneVision & $87.9\%$ & 1225.0 & 63.3 & 1436.7 & 275.0 & $95.2\%$ & $81.7\%$ & $88.7\%$ & $42.9\%$ \\
        14 & PaliGemma2-3B & $87.4\%$ & 1195.0 & 40.0 & 1460.0 & 305.0 & $96.8\%$ & $79.7\%$ & $88.5\%$ & $41.2\%$ \\
        15 & InternVL2.5-38B & $86.9\%$ & 1222.0 & 90.3 & 1409.7 & 278.0 & $93.3\%$ & $81.5\%$ & $87.7\%$ & $43.7\%$ \\
        16 & PaliGemma2-10B & $86.6\%$ & 1177.0 & 42.7 & 1457.3 & 323.0 & $96.5\%$ & $78.5\%$ & $87.8\%$ & $40.7\%$ \\
        17 & PaliGemma-3B & $85.7\%$ & 1160.0 & 46.0 & 1454.0 & 340.0 & $96.2\%$ & $77.3\%$ & $87.1\%$ & $40.2\%$ \\
        \bottomrule
    \end{tabular}
    \caption{\textbf{POPE - Mean}}
    \label{tab:POPE-Mean}\end{table*}

\begin{table*}[]
    \centering
    \begin{tabular}{rl|c|c|c|c|c|c|c|c|c}
        \toprule
        Pos. & Model & F1 & TP & FP & TN & FN & Precision & Recall & ACC & Yes Ratio\\
        \midrule
        1 & InternVL2.5-26B & $94.7\%$ & 1402 & 60 & 1440 & 98 & $95.9\%$ & $93.5\%$ & $94.7\%$ & $48.7\%$ \\
        2 & InternVL2.5-8B-MPO & $94.5\%$ & 1390 & 51 & 1449 & 110 & $96.5\%$ & $92.7\%$ & $94.6\%$ & $48.0\%$ \\
        3 & InternVL2.5-26B-MPO & $94.5\%$ & 1439 & 107 & 1393 & 61 & $93.1\%$ & $95.9\%$ & $94.4\%$ & $51.5\%$ \\
        4 & InternVL2.5-8B & $94.3\%$ & 1381 & 48 & 1452 & 119 & $96.6\%$ & $92.1\%$ & $94.4\%$ & $47.6\%$ \\
        5 & LLaVa-NeXT-Llama & $92.6\%$ & 1368 & 88 & 1412 & 132 & $94.0\%$ & $91.2\%$ & $92.7\%$ & $48.5\%$ \\
        6 & Ovis2-2B & $92.4\%$ & 1330 & 49 & 1451 & 170 & $96.4\%$ & $88.7\%$ & $92.7\%$ & $46.0\%$ \\
        7 & LLaVa-NeXT-Mistral & $92.0\%$ & 1335 & 67 & 1433 & 165 & $95.2\%$ & $89.0\%$ & $92.3\%$ & $46.7\%$ \\
        8 & Ovis2-4B & $91.8\%$ & 1294 & 24 & 1476 & 206 & $98.2\%$ & $86.3\%$ & $92.3\%$ & $43.9\%$ \\
        9 & Ovis2-8B & $91.7\%$ & 1290 & 22 & 1478 & 210 & $98.3\%$ & $86.0\%$ & $92.3\%$ & $43.7\%$ \\
        10 & Ovis2-1B & $91.7\%$ & 1314 & 53 & 1447 & 186 & $96.1\%$ & $87.6\%$ & $92.0\%$ & $45.6\%$ \\
        11 & LLaVa-NeXT-Vicuna & $91.6\%$ & 1318 & 60 & 1440 & 182 & $95.6\%$ & $87.9\%$ & $91.9\%$ & $45.9\%$ \\
        12 & InternVL2.5-78B & $90.4\%$ & 1246 & 12 & 1488 & 254 & $99.0\%$ & $83.1\%$ & $91.1\%$ & $41.9\%$ \\
        13 & LLaVA-OneVision & $89.3\%$ & 1225 & 20 & 1480 & 275 & $98.4\%$ & $81.7\%$ & $90.2\%$ & $41.5\%$ \\
        14 & InternVL2.5-38B & $89.1\%$ & 1222 & 20 & 1480 & 278 & $98.4\%$ & $81.5\%$ & $90.1\%$ & $41.4\%$ \\
        15 & PaliGemma2-3B & $88.4\%$ & 1195 & 10 & 1490 & 305 & $99.2\%$ & $79.7\%$ & $89.5\%$ & $40.2\%$ \\
        16 & PaliGemma2-10B & $87.5\%$ & 1177 & 12 & 1488 & 323 & $99.0\%$ & $78.5\%$ & $88.8\%$ & $39.6\%$ \\
        17 & PaliGemma-3B & $86.9\%$ & 1160 & 11 & 1489 & 340 & $99.1\%$ & $77.3\%$ & $88.3\%$ & $39.0\%$ \\
        \bottomrule
    \end{tabular}
    \caption{\textbf{POPE - Random}}
    \label{tab:POPE-random}\end{table*}

\begin{table*}[]
    \centering
    \begin{tabular}{rl|c|c|c|c|c|c|c|c|c}
        \toprule
        Pos. & Model & F1 & TP & FP & TN & FN & Precision & Recall & ACC & Yes Ratio\\
        \midrule
        1 & Ovis2-8B & $90.1\%$ & 1290 & 74 & 1426 & 210 & $94.6\%$ & $86.0\%$ & $90.5\%$ & $45.5\%$ \\
        2 & LLaVa-NeXT-Llama & $89.9\%$ & 1368 & 176 & 1324 & 132 & $88.6\%$ & $91.2\%$ & $89.7\%$ & $51.5\%$ \\
        3 & Ovis2-4B & $89.8\%$ & 1294 & 88 & 1412 & 206 & $93.6\%$ & $86.3\%$ & $90.2\%$ & $46.1\%$ \\
        4 & LLaVa-NeXT-Mistral & $89.7\%$ & 1335 & 142 & 1358 & 165 & $90.4\%$ & $89.0\%$ & $89.8\%$ & $49.2\%$ \\
        5 & InternVL2.5-26B & $89.6\%$ & 1402 & 229 & 1271 & 98 & $86.0\%$ & $93.5\%$ & $89.1\%$ & $54.4\%$ \\
        6 & Ovis2-2B & $89.2\%$ & 1330 & 153 & 1347 & 170 & $89.7\%$ & $88.7\%$ & $89.2\%$ & $49.4\%$ \\
        7 & InternVL2.5-8B & $89.1\%$ & 1381 & 218 & 1282 & 119 & $86.4\%$ & $92.1\%$ & $88.8\%$ & $53.3\%$ \\
        8 & LLaVa-NeXT-Vicuna & $89.0\%$ & 1318 & 144 & 1355 & 182 & $90.2\%$ & $87.9\%$ & $89.1\%$ & $48.7\%$ \\
        9 & InternVL2.5-78B & $88.6\%$ & 1246 & 68 & 1432 & 254 & $94.8\%$ & $83.1\%$ & $89.3\%$ & $43.8\%$ \\
        10 & InternVL2.5-8B-MPO & $88.5\%$ & 1390 & 250 & 1250 & 110 & $84.8\%$ & $92.7\%$ & $88.0\%$ & $54.7\%$ \\
        11 & Ovis2-1B & $88.4\%$ & 1314 & 160 & 1340 & 186 & $89.1\%$ & $87.6\%$ & $88.5\%$ & $49.1\%$ \\
        12 & InternVL2.5-26B-MPO & $88.3\%$ & 1439 & 319 & 1181 & 61 & $81.9\%$ & $95.9\%$ & $87.3\%$ & $58.6\%$ \\
        13 & LLaVA-OneVision & $87.8\%$ & 1225 & 67 & 1433 & 275 & $94.8\%$ & $81.7\%$ & $88.6\%$ & $43.1\%$ \\
        14 & PaliGemma2-3B & $87.2\%$ & 1195 & 46 & 1454 & 305 & $96.3\%$ & $79.7\%$ & $88.3\%$ & $41.4\%$ \\
        15 & InternVL2.5-38B & $86.7\%$ & 1222 & 97 & 1403 & 278 & $92.6\%$ & $81.5\%$ & $87.5\%$ & $44.0\%$ \\
        16 & PaliGemma2-10B & $86.7\%$ & 1177 & 39 & 1461 & 323 & $96.8\%$ & $78.5\%$ & $87.9\%$ & $40.5\%$ \\
        17 & PaliGemma-3B & $85.6\%$ & 1160 & 49 & 1451 & 340 & $95.9\%$ & $77.3\%$ & $87.0\%$ & $40.3\%$ \\
        \bottomrule
    \end{tabular}
    \caption{\textbf{POPE - Popular}}
    \label{tab:POPE-popular}\end{table*}

\begin{table*}[]
    \centering
    \begin{tabular}{rl|c|c|c|c|c|c|c|c|c}
        \toprule
        Pos. & Model & F1 & TP & FP & TN & FN & Precision & Recall & ACC & Yes Ratio\\
        \midrule
        1 & Ovis2-8B & $88.3\%$ & 1290 & 132 & 1368 & 210 & $90.7\%$ & $86.0\%$ & $88.6\%$ & $47.4\%$ \\
        2 & Ovis2-4B & $88.0\%$ & 1294 & 146 & 1354 & 206 & $89.9\%$ & $86.3\%$ & $88.3\%$ & $48.0\%$ \\
        3 & InternVL2.5-78B & $87.2\%$ & 1246 & 112 & 1388 & 254 & $91.8\%$ & $83.1\%$ & $87.8\%$ & $45.3\%$ \\
        4 & LLaVA-OneVision & $86.6\%$ & 1225 & 103 & 1397 & 275 & $92.2\%$ & $81.7\%$ & $87.4\%$ & $44.3\%$ \\
        5 & PaliGemma2-3B & $86.6\%$ & 1195 & 64 & 1436 & 305 & $94.9\%$ & $79.7\%$ & $87.7\%$ & $42.0\%$ \\
        6 & Ovis2-1B & $86.5\%$ & 1314 & 224 & 1276 & 186 & $85.4\%$ & $87.6\%$ & $86.3\%$ & $51.3\%$ \\
        7 & Ovis2-2B & $86.5\%$ & 1330 & 245 & 1255 & 170 & $84.4\%$ & $88.7\%$ & $86.2\%$ & $52.5\%$ \\
        8 & LLaVa-NeXT-Mistral & $86.3\%$ & 1335 & 258 & 1242 & 165 & $83.8\%$ & $89.0\%$ & $85.9\%$ & $53.1\%$ \\
        9 & InternVL2.5-26B & $85.9\%$ & 1402 & 361 & 1139 & 98 & $79.5\%$ & $93.5\%$ & $84.7\%$ & $58.8\%$ \\
        10 & LLaVa-NeXT-Vicuna & $85.8\%$ & 1318 & 254 & 1246 & 182 & $83.8\%$ & $87.9\%$ & $85.5\%$ & $52.4\%$ \\
        11 & LLaVa-NeXT-Llama & $85.6\%$ & 1368 & 328 & 1172 & 132 & $80.7\%$ & $91.2\%$ & $84.7\%$ & $56.5\%$ \\
        12 & InternVL2.5-8B & $85.6\%$ & 1381 & 346 & 1154 & 119 & $80.0\%$ & $92.1\%$ & $84.5\%$ & $57.6\%$ \\
        13 & PaliGemma2-10B & $85.5\%$ & 1177 & 77 & 1423 & 323 & $93.9\%$ & $78.5\%$ & $86.7\%$ & $41.8\%$ \\
        14 & InternVL2.5-8B-MPO & $85.2\%$ & 1390 & 371 & 1129 & 110 & $78.9\%$ & $92.7\%$ & $84.0\%$ & $58.7\%$ \\
        15 & InternVL2.5-38B & $85.0\%$ & 1222 & 154 & 1346 & 278 & $88.8\%$ & $81.5\%$ & $85.6\%$ & $45.9\%$ \\
        16 & PaliGemma-3B & $84.7\%$ & 1160 & 78 & 1422 & 340 & $93.7\%$ & $77.3\%$ & $86.1\%$ & $41.3\%$ \\
        17 & InternVL2.5-26B-MPO & $83.5\%$ & 1439 & 506 & 994 & 61 & $74.0\%$ & $95.9\%$ & $81.1\%$ & $64.8\%$ \\
        \bottomrule
    \end{tabular}
    \caption{\textbf{POPE - Adversarial}}
    \label{tab:POPE-adversarial}\end{table*}

\section{Appendix: Full Results RePOPE}
\label{app:full-repope}

We show all results for all models on the random (Tab.~\ref{tab:RePOPE-random}), popular (Tab.~\ref{tab:RePOPE-popular}), and adversarial (Tab.~\ref{tab:RePOPE-adversarial}) split as well as the mean over all three subsets (Tab.~\ref{tab:RePOPE-Mean}).

\begin{table*}[]
    \centering
    \begin{tabular}{rl|c|c|c|c|c|c|c|c|c}
        \toprule
        Pos. & Model & F1 & TP & FP & TN & FN & Precision & Recall & ACC & Yes Ratio\\
        \midrule
        1 & Ovis2-4B & $94.2\%$ & 1123.7 & 84.0 & 1464.7 & 56.0 & $93.1\%$ & $95.3\%$ & $94.8\%$ & $44.3\%$ \\
        2 & InternVL2.5-78B & $94.1\%$ & 1101.3 & 60.0 & 1488.7 & 78.3 & $94.9\%$ & $93.4\%$ & $94.9\%$ & $42.6\%$ \\
        3 & Ovis2-8B & $94.1\%$ & 1116.0 & 77.3 & 1471.3 & 63.7 & $93.6\%$ & $94.6\%$ & $94.8\%$ & $43.8\%$ \\
        4 & PaliGemma2-3B & $92.9\%$ & 1062.3 & 44.3 & 1504.3 & 117.3 & $96.0\%$ & $90.1\%$ & $94.1\%$ & $40.6\%$ \\
        5 & LLaVA-OneVision & $92.8\%$ & 1078.0 & 66.0 & 1482.7 & 101.7 & $94.3\%$ & $91.4\%$ & $93.8\%$ & $41.9\%$ \\
        6 & Ovis2-1B & $92.2\%$ & 1128.3 & 141.3 & 1407.3 & 51.3 & $89.0\%$ & $95.7\%$ & $92.9\%$ & $46.6\%$ \\
        7 & InternVL2.5-38B & $92.1\%$ & 1073.7 & 78.7 & 1470.0 & 106.0 & $93.3\%$ & $91.0\%$ & $93.2\%$ & $42.3\%$ \\
        8 & PaliGemma2-10B & $92.0\%$ & 1044.3 & 46.3 & 1502.3 & 135.3 & $95.8\%$ & $88.5\%$ & $93.3\%$ & $40.0\%$ \\
        9 & Ovis2-2B & $91.8\%$ & 1131.3 & 154.3 & 1394.3 & 48.3 & $88.2\%$ & $95.9\%$ & $92.5\%$ & $47.2\%$ \\
        10 & InternVL2.5-26B & $91.3\%$ & 1167.7 & 214.7 & 1334.0 & 12.0 & $84.8\%$ & $99.0\%$ & $91.6\%$ & $50.7\%$ \\
        11 & LLaVa-NeXT-Vicuna & $91.1\%$ & 1114.3 & 152.7 & 1395.7 & 65.3 & $88.1\%$ & $94.5\%$ & $92.0\%$ & $46.5\%$ \\
        12 & InternVL2.5-8B & $91.1\%$ & 1153.7 & 202.7 & 1346.0 & 26.0 & $85.4\%$ & $97.8\%$ & $91.6\%$ & $49.8\%$ \\
        13 & LLaVa-NeXT-Mistral & $91.1\%$ & 1120.0 & 161.3 & 1387.3 & 59.7 & $87.6\%$ & $94.9\%$ & $91.9\%$ & $47.0\%$ \\
        14 & PaliGemma-3B & $90.7\%$ & 1023.0 & 53.0 & 1495.7 & 156.7 & $95.1\%$ & $86.7\%$ & $92.3\%$ & $39.5\%$ \\
        15 & InternVL2.5-8B-MPO & $90.6\%$ & 1158.7 & 222.7 & 1326.0 & 21.0 & $84.3\%$ & $98.2\%$ & $91.0\%$ & $50.7\%$ \\
        16 & LLaVa-NeXT-Llama & $89.8\%$ & 1133.0 & 212.7 & 1336.0 & 46.7 & $84.4\%$ & $96.1\%$ & $90.5\%$ & $49.4\%$ \\
        17 & InternVL2.5-26B-MPO & $88.2\%$ & 1177.0 & 318.7 & 1230.0 & 2.7 & $79.2\%$ & $99.8\%$ & $88.2\%$ & $54.9\%$ \\
        \bottomrule
    \end{tabular}
    \caption{\textbf{RePOPE - Mean}}
    \label{tab:RePOPE-Mean}\end{table*}

\begin{table*}[]
    \centering
    \begin{tabular}{rl|c|c|c|c|c|c|c|c|c}
        \toprule
        Pos. & Model & F1 & TP & FP & TN & FN & Precision & Recall & ACC & Yes Ratio\\
        \midrule
        1 & Ovis2-4B & $95.9\%$ & 1113 & 49 & 1566 & 46 & $95.8\%$ & $96.0\%$ & $96.6\%$ & $41.9\%$ \\
        2 & Ovis2-8B & $95.9\%$ & 1109 & 45 & 1570 & 50 & $96.1\%$ & $95.7\%$ & $96.6\%$ & $41.6\%$ \\
        3 & InternVL2.5-78B & $95.5\%$ & 1090 & 33 & 1582 & 69 & $97.1\%$ & $94.0\%$ & $96.3\%$ & $40.5\%$ \\
        4 & InternVL2.5-26B & $95.0\%$ & 1150 & 111 & 1504 & 9 & $91.2\%$ & $99.2\%$ & $95.7\%$ & $45.5\%$ \\
        5 & InternVL2.5-8B-MPO & $94.9\%$ & 1142 & 105 & 1510 & 17 & $91.6\%$ & $98.5\%$ & $95.6\%$ & $45.0\%$ \\
        6 & InternVL2.5-8B & $94.9\%$ & 1138 & 101 & 1514 & 21 & $91.8\%$ & $98.2\%$ & $95.6\%$ & $44.7\%$ \\
        7 & Ovis2-1B & $94.8\%$ & 1118 & 81 & 1534 & 41 & $93.2\%$ & $96.5\%$ & $95.6\%$ & $43.2\%$ \\
        8 & Ovis2-2B & $94.7\%$ & 1121 & 87 & 1528 & 38 & $92.8\%$ & $96.7\%$ & $95.5\%$ & $43.5\%$ \\
        9 & LLaVA-OneVision & $94.3\%$ & 1072 & 42 & 1573 & 87 & $96.2\%$ & $92.5\%$ & $95.3\%$ & $40.2\%$ \\
        10 & PaliGemma2-3B & $94.2\%$ & 1057 & 29 & 1586 & 102 & $97.3\%$ & $91.2\%$ & $95.3\%$ & $39.1\%$ \\
        11 & InternVL2.5-38B & $93.9\%$ & 1063 & 41 & 1574 & 96 & $96.3\%$ & $91.7\%$ & $95.1\%$ & $39.8\%$ \\
        12 & LLaVa-NeXT-Vicuna & $93.6\%$ & 1104 & 96 & 1519 & 55 & $92.0\%$ & $95.3\%$ & $94.6\%$ & $43.3\%$ \\
        13 & PaliGemma2-10B & $93.4\%$ & 1041 & 29 & 1586 & 118 & $97.3\%$ & $89.8\%$ & $94.7\%$ & $38.6\%$ \\
        14 & LLaVa-NeXT-Mistral & $93.3\%$ & 1110 & 110 & 1505 & 49 & $91.0\%$ & $95.8\%$ & $94.3\%$ & $44.0\%$ \\
        15 & InternVL2.5-26B-MPO & $92.8\%$ & 1157 & 177 & 1438 & 2 & $86.7\%$ & $99.8\%$ & $93.5\%$ & $48.1\%$ \\
        16 & LLaVa-NeXT-Llama & $92.5\%$ & 1122 & 146 & 1469 & 37 & $88.5\%$ & $96.8\%$ & $93.4\%$ & $45.7\%$ \\
        17 & PaliGemma-3B & $92.1\%$ & 1018 & 34 & 1581 & 141 & $96.8\%$ & $87.8\%$ & $93.7\%$ & $37.9\%$ \\
        \bottomrule
    \end{tabular}
    \caption{\textbf{RePOPE - Random}}
    \label{tab:RePOPE-random}\end{table*}

\begin{table*}[]
    \centering
    \begin{tabular}{rl|c|c|c|c|c|c|c|c|c}
        \toprule
        Pos. & Model & F1 & TP & FP & TN & FN & Precision & Recall & ACC & Yes Ratio\\
        \midrule
        1 & Ovis2-4B & $94.1\%$ & 1131 & 80 & 1454 & 62 & $93.4\%$ & $94.8\%$ & $94.8\%$ & $44.4\%$ \\
        2 & Ovis2-8B & $93.9\%$ & 1121 & 73 & 1461 & 72 & $93.9\%$ & $94.0\%$ & $94.7\%$ & $43.8\%$ \\
        3 & InternVL2.5-78B & $93.7\%$ & 1106 & 62 & 1472 & 87 & $94.7\%$ & $92.7\%$ & $94.5\%$ & $42.8\%$ \\
        4 & LLaVA-OneVision & $92.6\%$ & 1083 & 64 & 1470 & 110 & $94.4\%$ & $90.8\%$ & $93.6\%$ & $42.1\%$ \\
        5 & PaliGemma2-3B & $92.4\%$ & 1065 & 47 & 1487 & 128 & $95.8\%$ & $89.3\%$ & $93.6\%$ & $40.8\%$ \\
        6 & InternVL2.5-38B & $91.8\%$ & 1080 & 80 & 1454 & 113 & $93.1\%$ & $90.5\%$ & $92.9\%$ & $42.5\%$ \\
        7 & Ovis2-2B & $91.7\%$ & 1139 & 152 & 1382 & 54 & $88.2\%$ & $95.5\%$ & $92.4\%$ & $47.3\%$ \\
        8 & PaliGemma2-10B & $91.7\%$ & 1046 & 43 & 1491 & 147 & $96.1\%$ & $87.7\%$ & $93.0\%$ & $39.9\%$ \\
        9 & Ovis2-1B & $91.6\%$ & 1135 & 149 & 1385 & 58 & $88.4\%$ & $95.1\%$ & $92.4\%$ & $47.1\%$ \\
        10 & LLaVa-NeXT-Vicuna & $91.4\%$ & 1121 & 139 & 1394 & 72 & $89.0\%$ & $94.0\%$ & $92.3\%$ & $46.2\%$ \\
        11 & LLaVa-NeXT-Mistral & $91.3\%$ & 1124 & 145 & 1389 & 69 & $88.6\%$ & $94.2\%$ & $92.2\%$ & $46.5\%$ \\
        12 & InternVL2.5-26B & $91.1\%$ & 1179 & 216 & 1318 & 14 & $84.5\%$ & $98.8\%$ & $91.6\%$ & $51.2\%$ \\
        13 & InternVL2.5-8B & $90.9\%$ & 1164 & 205 & 1329 & 29 & $85.0\%$ & $97.6\%$ & $91.4\%$ & $50.2\%$ \\
        14 & LLaVa-NeXT-Llama & $90.5\%$ & 1141 & 187 & 1347 & 52 & $85.9\%$ & $95.6\%$ & $91.2\%$ & $48.7\%$ \\
        15 & PaliGemma-3B & $90.2\%$ & 1025 & 54 & 1480 & 168 & $95.0\%$ & $85.9\%$ & $91.9\%$ & $39.6\%$ \\
        16 & InternVL2.5-8B-MPO & $90.1\%$ & 1170 & 235 & 1299 & 23 & $83.3\%$ & $98.1\%$ & $90.5\%$ & $51.5\%$ \\
        17 & InternVL2.5-26B-MPO & $88.2\%$ & 1190 & 315 & 1219 & 3 & $79.1\%$ & $99.7\%$ & $88.3\%$ & $55.2\%$ \\
        \bottomrule
    \end{tabular}
    \caption{\textbf{RePOPE - Popular}}
    \label{tab:RePOPE-popular}\end{table*}

\begin{table*}[]
    \centering
    \begin{tabular}{rl|c|c|c|c|c|c|c|c|c}
        \toprule
        Pos. & Model & F1 & TP & FP & TN & FN & Precision & Recall & ACC & Yes Ratio\\
        \midrule
        1 & InternVL2.5-78B & $93.1\%$ & 1108 & 85 & 1412 & 79 & $92.9\%$ & $93.3\%$ & $93.9\%$ & $44.4\%$ \\
        2 & Ovis2-4B & $92.5\%$ & 1127 & 123 & 1374 & 60 & $90.2\%$ & $94.9\%$ & $93.2\%$ & $46.6\%$ \\
        3 & Ovis2-8B & $92.4\%$ & 1118 & 114 & 1383 & 69 & $90.7\%$ & $94.2\%$ & $93.2\%$ & $45.9\%$ \\
        4 & PaliGemma2-3B & $92.2\%$ & 1065 & 57 & 1440 & 122 & $94.9\%$ & $89.7\%$ & $93.3\%$ & $41.8\%$ \\
        5 & LLaVA-OneVision & $91.5\%$ & 1079 & 92 & 1405 & 108 & $92.1\%$ & $90.9\%$ & $92.5\%$ & $43.6\%$ \\
        6 & PaliGemma2-10B & $91.0\%$ & 1046 & 67 & 1430 & 141 & $94.0\%$ & $88.1\%$ & $92.3\%$ & $41.5\%$ \\
        7 & InternVL2.5-38B & $90.6\%$ & 1078 & 115 & 1382 & 109 & $90.4\%$ & $90.8\%$ & $91.7\%$ & $44.4\%$ \\
        8 & Ovis2-1B & $90.1\%$ & 1132 & 194 & 1303 & 55 & $85.4\%$ & $95.4\%$ & $90.7\%$ & $49.4\%$ \\
        9 & PaliGemma-3B & $89.8\%$ & 1026 & 71 & 1426 & 161 & $93.5\%$ & $86.4\%$ & $91.4\%$ & $40.9\%$ \\
        10 & Ovis2-2B & $89.1\%$ & 1134 & 224 & 1273 & 53 & $83.5\%$ & $95.5\%$ & $89.7\%$ & $50.6\%$ \\
        11 & LLaVa-NeXT-Mistral & $88.6\%$ & 1126 & 229 & 1268 & 61 & $83.1\%$ & $94.9\%$ & $89.2\%$ & $50.5\%$ \\
        12 & LLaVa-NeXT-Vicuna & $88.4\%$ & 1118 & 223 & 1274 & 69 & $83.4\%$ & $94.2\%$ & $89.1\%$ & $50.0\%$ \\
        13 & InternVL2.5-26B & $87.7\%$ & 1174 & 317 & 1180 & 13 & $78.7\%$ & $98.9\%$ & $87.7\%$ & $55.6\%$ \\
        14 & InternVL2.5-8B & $87.5\%$ & 1159 & 302 & 1195 & 28 & $79.3\%$ & $97.6\%$ & $87.7\%$ & $54.4\%$ \\
        15 & InternVL2.5-8B-MPO & $86.9\%$ & 1164 & 328 & 1169 & 23 & $78.0\%$ & $98.1\%$ & $86.9\%$ & $55.6\%$ \\
        16 & LLaVa-NeXT-Llama & $86.5\%$ & 1136 & 305 & 1192 & 51 & $78.8\%$ & $95.7\%$ & $86.7\%$ & $53.7\%$ \\
        17 & InternVL2.5-26B-MPO & $83.5\%$ & 1184 & 464 & 1033 & 3 & $71.8\%$ & $99.7\%$ & $82.6\%$ & $61.4\%$ \\
        \bottomrule
    \end{tabular}
    \caption{\textbf{RePOPE - Adversarial}}
    \label{tab:RePOPE-adversarial}\end{table*}

\end{document}